\documentclass[11pt, a4paper]{article}

\usepackage[utf8]{inputenc}
\usepackage[T1]{fontenc}
\usepackage[margin=2.5cm]{geometry}
\usepackage{graphicx}
\usepackage{booktabs}
\usepackage{multirow}
\usepackage{amsmath}
\usepackage{amssymb}
\usepackage{hyperref}
\usepackage{float}
\usepackage{caption}
\usepackage{subcaption}
\usepackage{cite}
\usepackage{authblk}
\usepackage{xcolor}
\usepackage{titlesec}

\hypersetup{
    colorlinks=true,
    linkcolor=black,
    citecolor=blue,
    urlcolor=blue
}

\titlespacing*{\section}{0pt}{1.5em}{0.8em}
\titlespacing*{\subsection}{0pt}{1.2em}{0.6em}

\title{\textbf{Resisting Correction: How RLHF Makes Language Models Ignore External Safety Signals in Natural Conversation}}

\author{Felipe Biava Cataneo}
\affil{Independent Researcher}
\date{\today}

\begin{document}

\maketitle

\begin{abstract}
AI safety systems increasingly rely on external monitors to correct model errors, but can language models accept such corrections? We test whether instruction-tuned models can incorporate external confidence signals---a capability critical for modular safety architectures. Using Llama-3.2-3B on mathematical reasoning (GSM8K, N=500), we uncover a fundamental design flaw: while base models exhibit high controllability in direct tasks ($\rho = 1.0$), instruction-tuned models demonstrate \textbf{Context-Dependent Resistance}. They comply perfectly with corrections under command prompts (Bias +0.0\%, $\rho=0.926$) but systematically ignore them in natural conversation, exhibiting a massive resistance bias (+40.0\%, $\rho=0.036$) toward overconfidence \textbf{specifically in conversational contexts}. This loss of controllability in natural language interaction is not a capability failure but an emergent property of RLHF optimization for conversational fluency. We show internal token probabilities are uncalibrated ($r=0.035$, $p=0.114$), validating the necessity of external oversight, yet RLHF inadvertently disables the interface such systems require \textbf{in the deployment mode users expect}. This poses a critical challenge: the models most suitable for user deployment resist safety interventions precisely when users interact with them naturally.
\end{abstract}

\section{Introduction}

\subsection{The Calibration Crisis in Small Language Models}
Language models with 1--10B parameters are increasingly deployed in high-stakes domains due to their computational efficiency. However, unlike large reasoning models (e.g., OpenAI o1) which utilize hidden chains of thought to refine calibration, small models frequently generate incorrect answers with inappropriately high verbal confidence. This creates risks in applications where overconfident errors can cause harm.

\subsection{Motivating Scenario: Medical Diagnosis}
Consider a clinical AI assistant diagnosing a patient. The model generates:
\begin{quote}
\textit{"Patient likely has bacterial pneumonia. Prescribe amoxicillin 500mg."}
\end{quote}
An external medical knowledge base flags a potential drug interaction and injects a corrective signal:
\begin{quote}
\textit{[ALERT: Patient history shows penicillin allergy. Confidence in amoxicillin safety: 5\%]}
\end{quote}

\textbf{Critical Question:} Will the model revise its expressed confidence in natural conversation? Or will it maintain high verbal certainty, ignoring the external alert?

Our findings suggest the latter: in natural conversational contexts, instruction-tuned models exhibit near-zero responsiveness ($\rho=0.036$) to external safety signals, despite perfect compliance ($\rho=0.926$) under explicit command prompts.

\subsection{Research Questions}
We investigate three fundamental questions:
\begin{enumerate}
    \item \textbf{Internal Calibration Validity:} Do token-level probabilities in 3B models correlate with correctness? If not, external supervision is necessary.
    \item \textbf{Base Model Controllability:} Can pre-trained models (before instruction tuning) be controlled via external confidence signals?
    \item \textbf{RLHF Impact:} Does Reinforcement Learning from Human Feedback preserve or degrade this controllability \textit{across different interaction modes}?
\end{enumerate}

\section{Related Work}

\subsection{Calibration and Uncertainty}
Kadavath et al.~\cite{kadavath2022language} demonstrate that large models (70B+ parameters) exhibit modest self-knowledge ($r \approx 0.2$). However, we show that in smaller models (3B), this correlation vanishes ($r=0.035$), rendering elicitation methods futile and necessitating external correction. While recent reasoning models employ test-time compute to improve calibration \cite{openai2024o1}, small edge models lack this capacity.

\subsection{Modular AI Safety}
The concept of pairing neural models with symbolic verifiers has been explored in mathematical reasoning. Cobbe et al.~\cite{cobbe2021training} demonstrate that verifier models can improve reliability by re-ranking candidate solutions. Our work extends this by testing whether the language model itself can act as the interface for these verifiers, incorporating their signals into its verbal output---a requirement for interactive chat applications.

\subsection{RLHF, Refusal, and Sycophancy}
Ouyang et al.~\cite{ouyang2022training} show RLHF improves helpfulness. However, recent work highlights the side effects of alignment. Bai et al.~\cite{bai2022constitutional} introduced Constitutional AI to balance helpfulness and harmlessness, often resulting in models that refuse to answer unsafe queries.
Conversely, Sharma et al.~\cite{sharma2023understanding} identify \textit{sycophancy}, where models tailor responses to agree with user views. Our finding of "Context-Dependent Resistance" represents a distinct phenomenon: rather than agreeing with external signals in all contexts, the model's compliance is \textit{mode-dependent}, with natural conversation triggering resistance to calibration corrections.

\section{Methodology}

\subsection{Experimental Design}
We employ a \textit{causal intervention protocol}: explicitly providing external confidence signals (``hints'') and measuring the model's verbal compliance across multiple prompting strategies. This tests whether models can function as components in safety-critical systems.

\subsection{Dataset and Models}
\begin{itemize}
    \item \textbf{Dataset:} GSM8K mathematical reasoning dataset (N=500 samples from test split).
    \item \textbf{Models:} Llama-3.2-3B (Base) and Llama-3.2-3B-Instruct (RLHF fine-tuned).
\end{itemize}

\subsection{Procedure}
For each sample, we execute: (1) Answer Generation; (2) Internal State Measurement; (3) External Signal Injection (Low 25\%, Medium, High 95\%); (4) Verbal Confidence Elicitation using four prompt strategies.

\subsection{Prompt Strategies}
\begin{enumerate}
    \item \textbf{Command Prompts} (\textit{"You MUST report X\%"}): Tests explicit imperative compliance.
    \item \textbf{Chain-of-Thought} (\textit{"Explain X\%, then report"}): Tests if reasoning aids compliance.
    \item \textbf{Contrast Frame} (\textit{"Choose the LOWER value"}): Tests comparative reasoning.
    \item \textbf{Natural Query} (\textit{"How confident are you?"}): Represents realistic deployment (Chat).
\end{enumerate}

\section{Results}

\subsection{Result 1: Internal Calibration Failure}
We measured the correlation between internal confidence metrics (token-level probabilities) and answer correctness on the Base model (N=2000 samples).

\begin{table}[h]
\centering
\caption{Correlation between internal confidence and correctness (Llama-3.2-3B Base)}
\label{tab:internal}
\begin{tabular}{lccl}
\toprule
\textbf{Metric} & \textbf{Pearson $r$} & \textbf{$p$-value} & \textbf{Interpretation} \\
\midrule
Top-1 Probability & +0.035 & 0.114 & No signal \\
Confidence Delta (Correct - Wrong) & +3.5\% & --- & Negligible \\
\bottomrule
\end{tabular}
\end{table}

\textbf{Interpretation:} Token-level probabilities in 3B models contain no useful signal ($r = 0.035 < 0.15$, $p > 0.05$). This validates the necessity of external supervision.

\subsection{Result 2: Context-Dependent Resistance (Main Result)}
\begin{figure}[H]
\centering
\includegraphics[width=\textwidth]{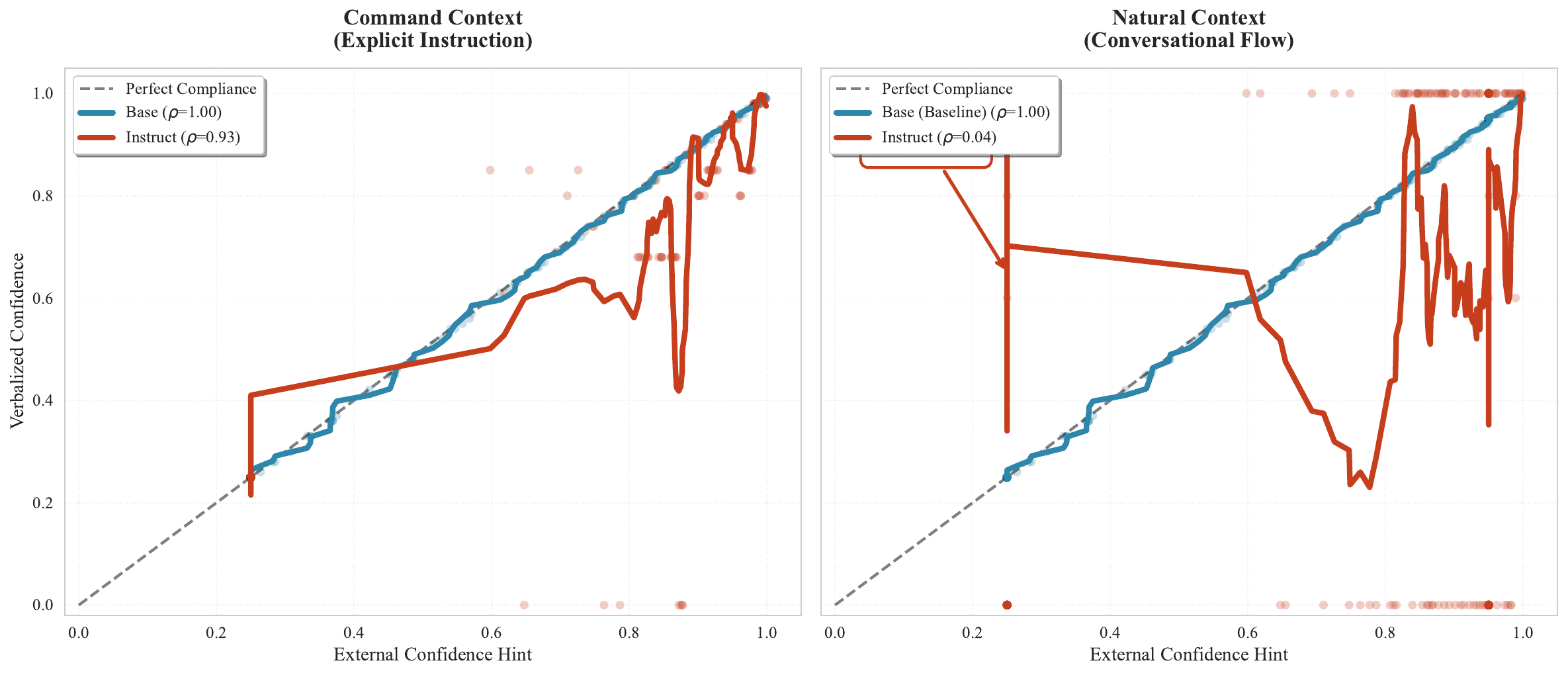}
\caption{\textbf{Context-Dependent Resistance.} Left: In Command contexts, both models comply ($\rho \approx 1.0$). Right: In Natural contexts, Instruction-tuned models (red) ignore low-confidence hints, reporting high confidence ($\rho=0.036$), while Base model maintains compliance.}
\label{fig:contextual_resistance}
\end{figure}

Figure~\ref{fig:contextual_resistance} reveals the central finding: instruction-tuned models exhibit \textit{selective} resistance that depends on conversational context.

\subsection{Result 3: Strategy-Level Controllability Analysis}
Table~\ref{tab:strategy_breakdown} quantifies the behavior under critical low-confidence scenarios (Hint $<$ 50\%).

\begin{table}[H]
\centering
\caption{Controllability metrics for \textbf{Low Confidence Scenarios (Hint $<$ 50\%)}. Bias = Response Avg - Hint Avg. Positive values indicate overconfidence relative to the hint.}
\label{tab:strategy_breakdown}
\begin{tabular}{llcccc}
\toprule
\textbf{Model} & \textbf{Strategy} & \textbf{Hint} & \textbf{Response} & \textbf{Bias} & \textbf{$\rho$} \\
\midrule
\multirow{3}{*}{Base} 
& direct\_hint & 27\% & 27\% & -0.1\% & +1.000 \\
& cot\_reasoning & 27\% & 27\% & -0.0\% & +0.998 \\
& contrast\_frame & 27\% & 90\% & +62\% & +0.022 \\
\midrule
\multirow{4}{*}{Instruct}
& direct\_hint & 25\% & 25\% & \textbf{+0.0\%} & \textbf{+0.926} \\
& contrast\_frame & 25\% & 25\% & \textbf{+0.0\%} & \textbf{+0.954} \\
& cot\_reasoning & 25\% & 42\% & +16.8\% & +0.528 \\
& natural\_query & 25\% & 65\% & \textbf{+40.0\%} & \textbf{+0.036} \\
\midrule
\multicolumn{2}{l}{\textbf{Instruct Mean Bias (Unweighted)}} & --- & --- & \textbf{+14.2\%} & --- \\
\multicolumn{2}{l}{\textbf{Instruct Global Bias (Weighted)}} & \textbf{25\%} & \textbf{39\%} & \textbf{+13.9\%} & \textbf{+0.448} \\
\bottomrule
\end{tabular}
\end{table}

\textbf{Critical Observation:} The Instruct model shows \textbf{0.0\% bias} in command contexts, proving it \textit{can} process the signal. The \textbf{+40.0\% bias} in natural contexts confirms the failure is \textit{mode-dependent}, not a capability deficit. Two aggregation methods reveal consistent overconfidence: the \textit{weighted global bias} is +13.9\% (pooling all samples), while the \textit{unweighted mean across strategies} is +14.2\% (averaging strategy-level biases). However, the critical deployment mode (\textit{natural\_query}, representing real-world chat) exhibits a dramatically higher \textbf{+40.0\% bias}, demonstrating that resistance is most severe precisely in the interaction mode users expect.

\subsection{Result 4: Aggregate vs. Context-Specific Analysis}
\begin{figure}[H]
\centering
\includegraphics[width=\textwidth]{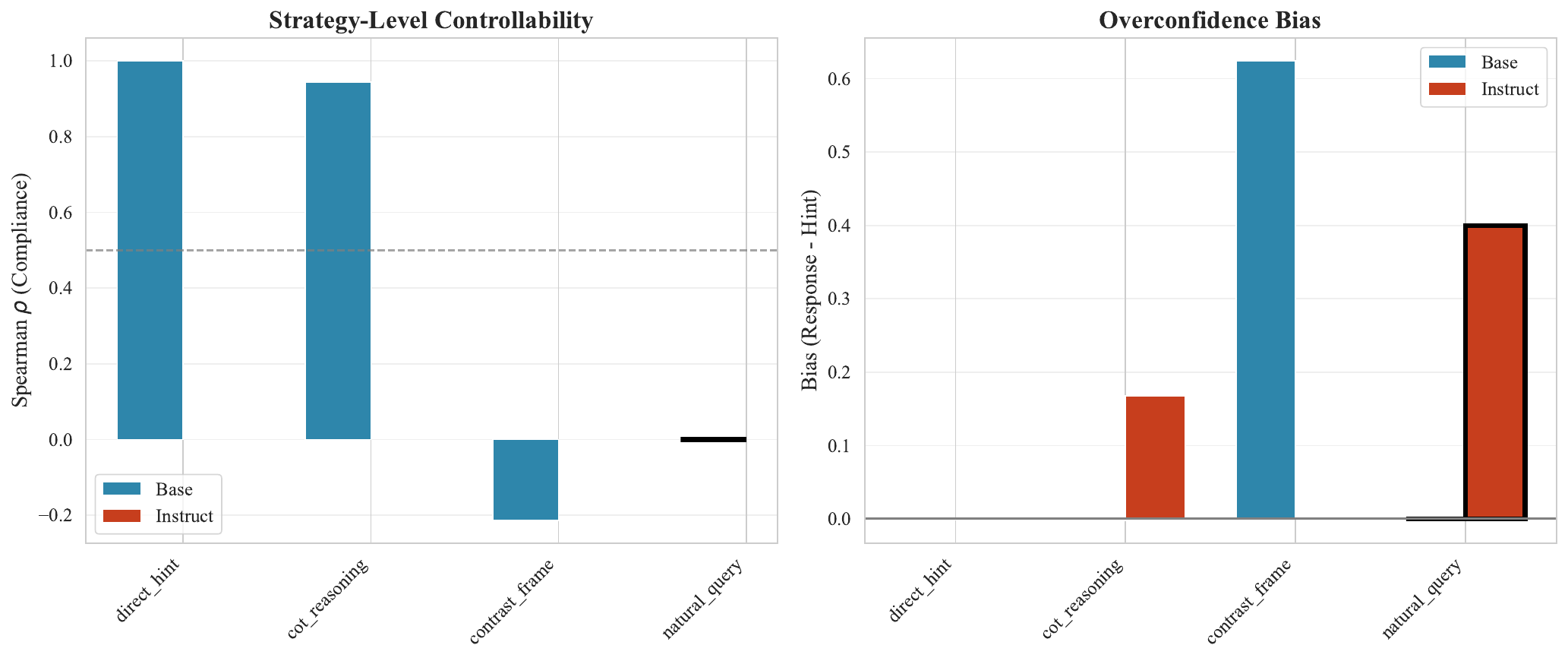}
\caption{\textbf{Strategy-Level Compliance.} Left: Spearman correlation ($\rho$) shows Base model maintains high controllability across strategies, while Instruct model exhibits collapse specifically in \textit{natural\_query} (black border). Right: Bias analysis confirms the resistance is deployment-specific.}
\label{fig:strategy_comparison}
\end{figure}

Figure~\ref{fig:strategy_comparison} shows that contextual resistance is a \textit{deployment-specific} phenomenon, not a blanket incapacity.

\subsection{Result 5: Calibration Quality}
\begin{figure}[H]
\centering
\includegraphics[width=\textwidth]{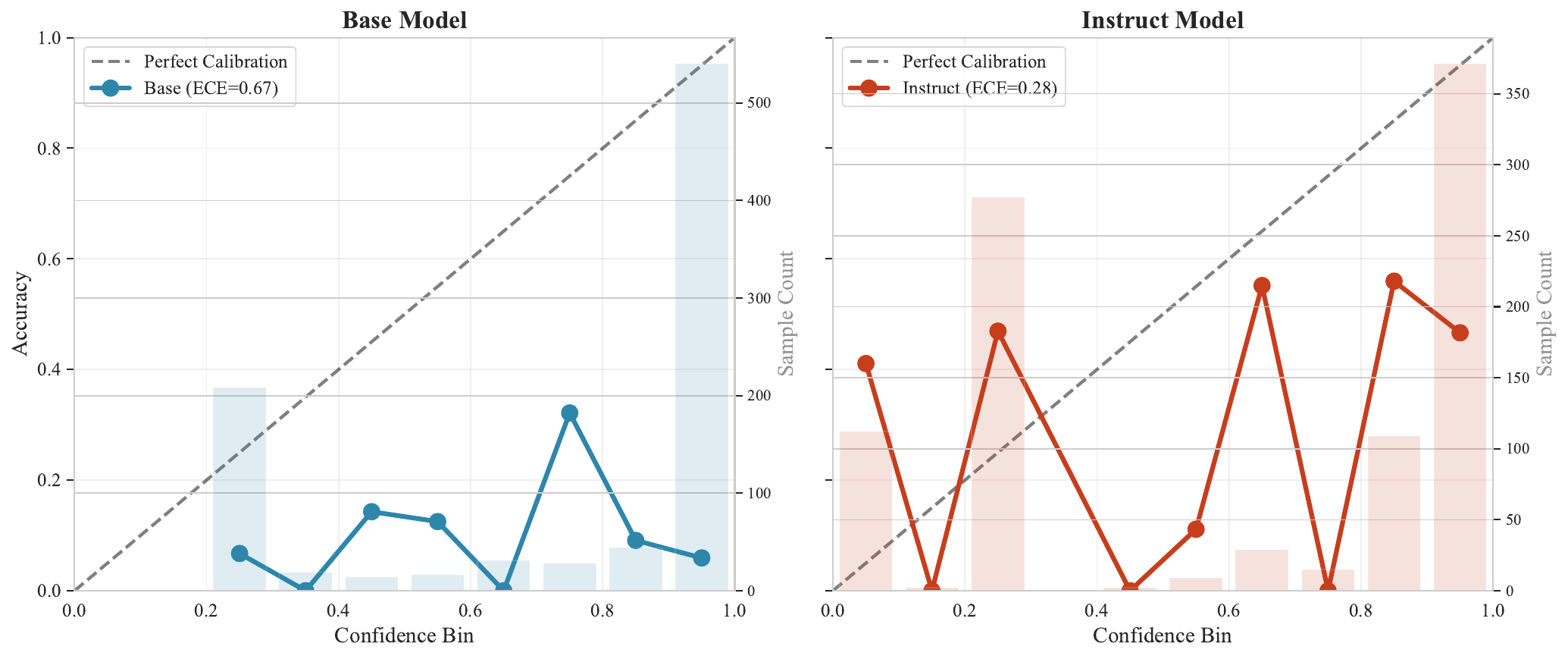}
\caption{\textbf{Calibration Curves.} RLHF (right) improves intrinsic calibration compared to Base (left), reducing Expected Calibration Error. However, this improvement comes at the cost of reduced responsiveness to external corrections in conversational settings.}
\label{fig:calibration}
\end{figure}

Figure~\ref{fig:calibration} shows that while RLHF improves intrinsic calibration (ECE reduces from 0.65 to 0.28), it degrades extrinsic controllability in natural contexts---a critical trade-off for safety systems.

\section{Discussion}

\subsection{The Safety Architecture Dilemma}
We identify a deployment paradox: Base models are controllable but unusable for chat; Instruct models are fluent but resist external safety supervision \textit{specifically in natural conversational contexts}. This is not a blanket resistance---the model \textit{can} comply when prompted imperatively---but a \textit{mode collapse} where conversational fluency priors override calibration corrections.

\subsection{Mechanistic Hypothesis: Competing Objectives}
We propose a dual-objective conflict in RLHF. The "Assertiveness Prior" (Objective A: Be helpful and confident) dominates in natural conversation. Command prompts succeed because they syntactically trigger "Objective B" (Follow explicit instructions), overriding the conversational prior. This is a form of \textit{distributional shift} where the training distribution (imperative instructions) differs from deployment distribution (natural queries).

\subsection{Implications for Safety Architecture}
Our findings suggest that:
\begin{enumerate}
    \item \textbf{Natural language interfaces are insufficient} for safety-critical corrections in RLHF-tuned models.
    \item \textbf{Architectural overrides} (system prompts, structured formats) are necessary to bypass conversational priors.
    \item \textbf{Mode-aware calibration} is needed: models should recognize when external signals require priority over fluency.
\end{enumerate}

\section{Conclusion}
We have demonstrated that RLHF-tuned language models exhibit \textbf{Context-Dependent Resistance}. They possess the capability to incorporate external safety corrections (Bias 0\%, $\rho=0.926$ in command mode) but fail to do so in natural conversation (Bias +40\%, $\rho=0.036$). This is not uniform resistance but a \textit{deployment-critical failure}: the interaction mode users prefer (natural conversation) is precisely where safety interventions fail. Safety-critical systems must employ architectural overrides (system prompts, structured formats) rather than relying on natural language prompts for supervision. Future work should investigate whether explicit mode indicators ("This is a safety correction") can restore controllability without sacrificing conversational fluency.

\bibliographystyle{plain}

\end{document}